*Title / name of your software*

TorchPRISM: **Pr**incipal **I**mage **S**ections **M**apping, a novel method for Convolutional Neural Network features visualization

*Authors / main developers*

Tomasz Szandała

*Abstract.*

In this paper we introduce a tool called Principal Image Sections Mapping - PRISM, dedicated for PyTorch, but can be easily ported to other deep learning frameworks. Presented software relies on Principal Component Analysis to visualize the most significant features recognized by a given Convolutional Neural Network. Moreover, it allows to display comparative set features between images processed in the same batch, therefore PRISM can be a method well synerging with technique Explanation by Example.

*Keywords:*

*1 Introduction*

Convolutional Neural Networks (CNNs) have enabled unprecedented breakthroughs in a variety of computer vision tasks like image classifica-tion [2], object detection [3,6], semantic segmenta-tion [4], image captioning [5] and visual question answering [7]. However, even with such unprecedented advancements, the lack of full understanding regarding the decisions made by deep learning models and absence of control over their internal processes act as major setbacks in critical decision-making processes, such as precision medicine or autonomous cars. In response, efforts are being made to make deep learning interpretable and controllable by humans.

A range of visualisation and explanation methods have been proposed. They have been valued against multiple factors, like speed, complexity, fidelity, contrastivity, sparsity and readability[1, 8, 9]. Among these methods we have identified a gap that relates to method Explanation by Example[10]. This technique allows to find images the most similar to a given image and therefore reason the features that lead to certain classification. However man can only assume the differences, but what if we could obtain a set of features present on a group of images?

We are introducing a new method that relies on Principal Component Analysis of features detected by neural network models and interpolates them on the initial image. The result of the formula is an RGB colored image mask that assigns one color to each feature identified by the model, see figure 1. Moreover the same color will be used to highlight the same feature across all pictures processed in the same batch, of course only if it is present on other samples. This allows to expose a comparative set of features between images processed in the same batch and thus straighten out technique Explanation by Example.

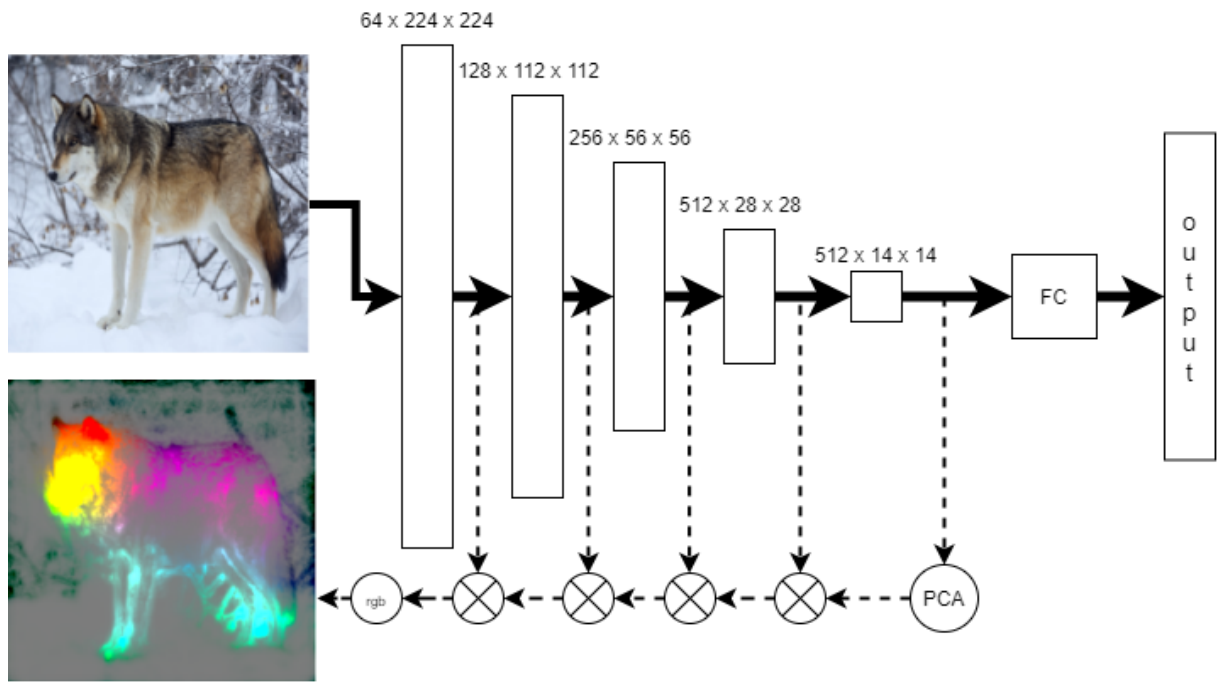

*Fig. 1. Conceptual diagram for obtaining PRISM map. Diagram displays a simplified schema of VGG-16 network.*

*2 Problems and Background*

*One of the main issues with deep networks is that they provide no visual output by themselves, that is, it is not possible to know what part of the image influenced the decision the most. As a consequence, there is a demand for methods that can help visualise or explain the decision-making process of such networks and make them understandable for humans.*

*There are multiple methods for inspecting models' attention. The most straightforward one and the oldest is the occlusion analysis. We hide a small part of the image and study the output. The more it has changed from original, the more significant the given area was. The most notable setback is the speed of the method. We have to analyze the image multiple times, once for each occluded region. This second limitation is that it does not take into account interdependencies between regions.*

*The next group of noteworthy methods rely on gradient backpropagation. Here we compute the gradient of the model's loss with respect to the weights. The gradient is a vector that contains a value for each weight, reflecting how much a small change in that weight will affect the output, essentially telling us which weights are most important for the loss. This single step through the network gives us an importance value for each pixel, which we display in the form of a heatmap. The enhancement of the method, guided backpropagation, blocks the backward flow of gradients from neurons whenever the output is negative, leaving only those gradients that result in increased output, which ultimately results in a less noisy interpretation. Since it only requires one pass through the network, produces a cleaner output, especially around the edges of the object it became preferred over occlusion.*

*However, a major issue with gradient backpropagation is that they don't work well when there are more than one class objects present in the image. To solve these issues a novel Class Activation Maps and its enhancements [11,12,13] were introduced. Here the authors found that the quality of interpretations improved when the gradients were taken at each filter of the last convolutional layer, instead of at the class score. The result is faster than occlusion and better than gradient backpropagation as it is targeted, therefore highlighting the most significant areas.*

*Apart from these a multitude of other methods were described, all with their own pros and cons and sometimes dedicated for limited application. Our attention caught method Explanation by Examples, introduced by Chen et al. in 2019 [10] and rated as the most preferred explanation styles according to the average non-technical end-user[1]. In input domains spanning visual, audio, and sensory data, explanation by nearest training examples offer users an opportunity to compare features across a test input and similarly mapped ground-truth examples. With the PRISM the discriminative feature are highlighted and therefore easier to spot.*

*3 Software Framework*

*3.1 Algorithm*

*Mathematics behind the tool relied on matrix computation. First we calculate activations for each subsequent layer for each image in the batch. In our case we just collect excitations during processing the batch and store them in an array $\mathbb{A}=<A_1,A_2,A_3,...,A_{|\mathbb{A}|}>$. Note that each A are activations for the entire batch.*

*Next we have to perform Principal Component Analysis for the activations of the last convolutional layer, $A_{|\mathbb{A}|}$. Initially $A_{|\mathbb{A}|}$ has the shape of: $n_b$ x c x h x w, which is: batch, channels, height and width. In PyTorch we have a similar operation for dimensionality reduction that is based on Singular Value Decomposition. To calculate SVD we have to reshape to the two dimensional matrix (A') and center it by subtracting the mean. In this way we treat each per-pixel channel response as a separate observation. Then we perform actual SVD to obtain 3 matrices: U is a unitary matrix, S is the diagonal matrix of singular values and V is a matrix of eigenvectors (each column is an eigenvector). Multiplication of U and S gives us the desired Principal Component Analysis of the final layer (A''). Moreover we are interested only in the 3 most significant vectors, according to PCA, therefore we are clipping only 3 first columns of $A_{PCA}$. This will result in easy visual interpretation of the outcome, since we can now consider each column as a respective channel: red, green and blue.*

$$A_{|\mathbb{A}|}^{n \times c \times h \times w} \xrightarrow{reshape} A_{|\mathbb{A}|}^{v \times c} = A'$$
$$A'' = A' - mean(A')$$
$$U \times S \times V^T = svd(A'')$$
$$A_{PCA} = U \times S$$

*Since we will be working on PyTorch shaped tensors we need to restore its shape to the original one. Currently we have obtained principal features mapping, which now has to be upsampled to the original image. For this purpose we used Torch's interpolation function in bilinear mode. Also to sharpen each feature visualisation we multiply obtained interpolation matrix with matrix of sums for each channel in the given layer.*

$$\forall_{A_i} \in \mathbb{A}: A'_{PCA} = interpolation(\ shape(A_{PCA}) \to shape(A_i)\ ) \times \sum_c A_i^{c \times h_i \times w_i}$$

*Finally, we need to convert the ultimate matrix to RGB colorspace, so it can be displayed as an image. It is just clipping to <-1,1>, for better color discrimination, and normalize to <0,1>. Concluding image has each color corresponding to a single feature identified by network, however only in this given batch. In other words, obtained features staining is specific to this one set of images.*

*3.2 Software Architecture*

*Module consists of actually one file. It contains one static class that uses a singleton-like-approach. It is enforced by PyTorch's hook behaviour, where we cannot refer to an external object, but we have to rely on global variables, or, like in this case, on static class attributes.*

*The class consists of:*

- *2 hidden variables to keep track of hooks catching excitation propagations*
- *several hidden methods used internally for PRISM computation*
- *4 public methods:*
  - *one for registering hooks into the model, one for disabling*
  - *one for pruning collected excitations*
  - *one for obtaining a PRISM tensor. Note that it removed saved excitations.*

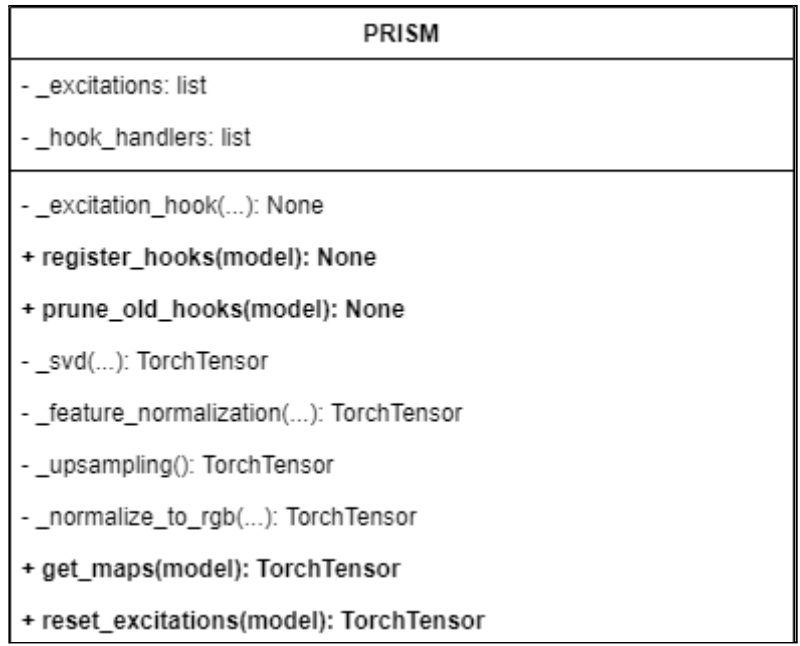

Fig. 2. UML class diagram for class PRISM

The PRISM class diagram has been pictured on the figure 2 above. The interface for private methods has been obscured.

3.3 Sample code snippet

Minimal working code can be found on listing 1. First we have to import PRISM and torch models., as well as functions for preparing input images as simple torch batch and function to draw batches. Next we have to load the model, in this case a pretrained vgg16 has been chosen and then we have to call the first PRISM method to register required hooks in the model.

```
from PRISM import PRISM
from torchvision import models
from my_utils import load_images, draw_input_n_prism

input_batch = load_images()

model = models.vgg16(pretrained=True)
model.eval()
PRISM.register_hooks(model)

model(input_batch)
prism_maps_batch = PRISM.get_maps()

drawable_input = input_batch.permute(0, 2, 3, 1).detach().cpu().numpy()
```

```
drawable_prism_maps = prism_maps_batch.permute(0, 2, 3,
1).detach().cpu().numpy()
draw_input_n_prism(drawable_input_batch, drawable_prism_maps_batch)
```

With such a prepared model we can perform the classification and, since the actual output is not needed, we can just ignore it. Model execution is followed by using the second PRISM method to calculate features maps for the processed batch. Finally we have to prepare both input and PRISM output so they can be drawn and as the last step we call a method that displays them using e.g. matplotlib.

*4 Empirical Results*

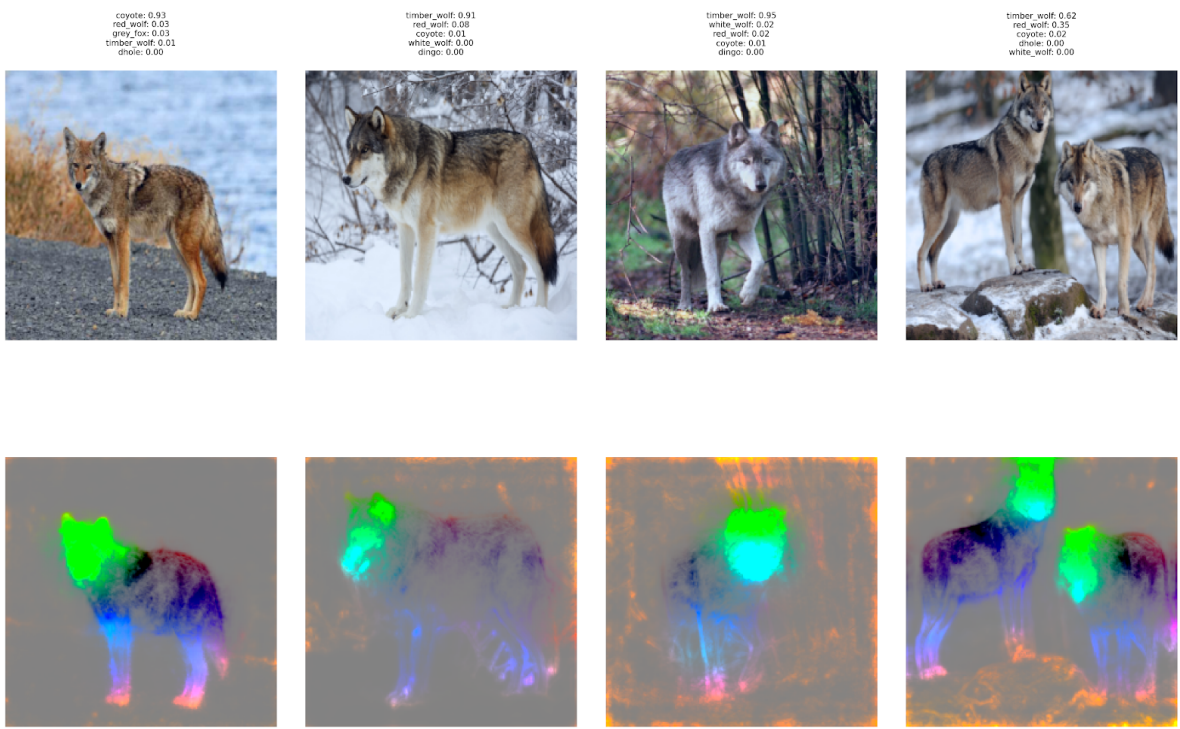

Fig. 3. PRISM output for several images of wolves and one with a coyote

In figure 3 we can see the result of PRISM for a batch of 3 wolf and one coyote images. We have picked up these pictures because timber wolves and coyotes were often confused with each other in our separate research. According to the network and therefore PRISM output we can assume that the main difference between wolf and coyote can be the mouth of these animals- here represented by a stain of aquamarine colour.
Furthermore we can see that some elements like pawns and legs are treated separately by the model. Noteworthy is also the network's attention as it clearly ignores some of the elements, like a trunk between wolves couple of some branches on other images.

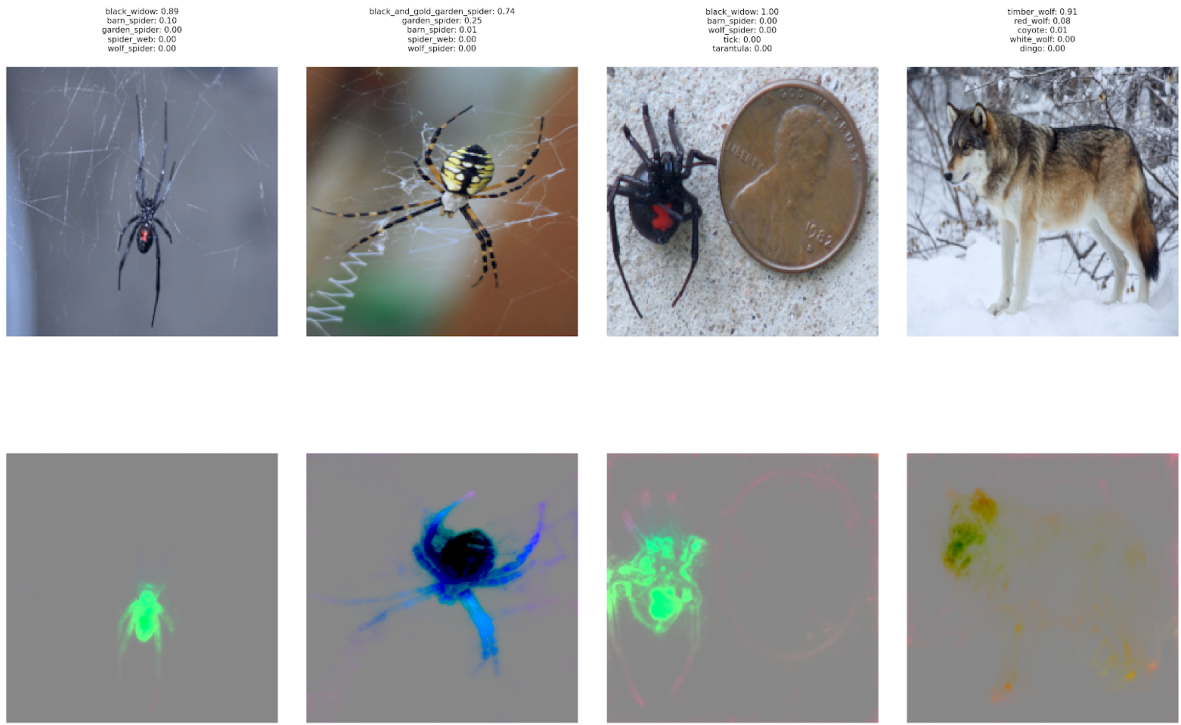

*Fig. 4. UML class diagram for class PRISM*

In the next figure, 4, we see the output of VGG-16 for 4 images, belonging to 3 different classes. We see that the network appears to treat black widow as a singular feature as on the first and third image it has detected the same features. Noticeable is that the model does not distinct red dots on a black-widow on the underside of their abdomen, which is considered to be their most discriminative feature by biologists.

## 6 Conclusions

Proposed tool allows users to determine which features were identified by the network on an image and contributed the most for the classification. Furthermore, the provided solution allows to compare features detected by the model across a set of images, therefore serves as an auxiliary interpreting method to Explanation by Example technique.